\documentclass[nonacm]{acmart}
\AtBeginDocument{%
  }

\usepackage{multirow} 

\begin{document}

\title{Half-Layered Neural Networks}

\author{Ethem Alpayd{\i}n}
\email{ethem.alpaydin@ozyegin.edu.tr}
\orcid{0000-0001-7506-0321}
\affiliation{%
  \institution{Department of Computer Engineering, {\"O}zye{\u g}in University}
  \city{{\c C}ekmek{\"o}y, TR-34794, {\.I}stanbul}
  \country{Turkey}
}

\renewcommand{\shortauthors}{Alpayd{\i}n}

\begin{abstract}
We propose a ``half'' layer of hidden units that has some of its weights randomly set and some of them trained. A half unit is composed of two stages: First, it takes a weighted sum of its inputs with fixed random weights, and second, the total activation is multiplied and then translated using two modifiable weights, before the result is passed through a nonlinearity. The number of modifiable weights of each hidden unit is thus two and does not depend on the fan-in. We show how such half units can be used in the first or any later layer in a deep network, possibly following convolutional layers. Our experiments on MNIST and FashionMNIST data sets indicate the promise of half layers, where we can achieve reasonable accuracy with a reduced number of parameters due to the regularizing effect of the randomized connections. 
\end{abstract}


\keywords{Do, Not, Us, This, Code, Put, the, Correct, Terms, for,
  Your, Paper}



\maketitle

\def\br{\mathbf{r}}
\def\bx{\mathbf{x}}
\def\bz{\mathbf{z}}
\def\cX{\mathcal X}

\section{Introduction}

We start with the multivariate linear model with $d$ inputs and $K$ outputs
\begin{equation}
y^t_i = f\left( \sum_{j=0}^d w_{ij}x^t_j \right), i=1,\ldots,K
\label{eq-linmod}
\end{equation}

\noindent with $x^t_0 = +1, \forall t$. The weights, $w_{ij}, i=1,\ldots,K, j=0,\ldots, d$, are trained on a training data set $\cX=\{ (\bx^t, \br^t)\}_{t=1}^N$ to minimize the total error: 
\begin{equation}
E(\{ w_{ij} \} ; \cX ]) = \sum_t \sum_i L(r^t_i, y^t_i)
\label{eq-linmod-err}
\end{equation}

\noindent where $L()$ is the loss that measures the discrepancy between the predicted $y^t_i$ for input $\bx^t$ and the desired $r^t_i$.

This model has several properties that make it attractive:
\begin{itemize}
\item {\em It is a general model:} It can be used for both numeric and logistic regression. In the former case, $f()$ is the identity function, $y^t_i, r^t_i\in\Re$, and we minimize the squared difference: $L(r^t_i,y^t_i)=(r^t_i-y^t_i)^2$. In the case of logistic regression, $f()$ is the logistic (sigmoid) or softmax function depending on whether there are two or more classes, respectively, $y^t_i\in [0,1], r^t_i\in\{0,1\}$, and we minimize the cross-entropy: $L(r^t_i, y^t_i)=r^t_i\log y^t_i$. 

\item {\em It can be trained optimally:} In the case of regression, the error function is quadratic and there is an analytical solution; in the case of logistic regression, the error function is convex and the local minimum we find (e.g., using gradient descent) is the global minimum. 

\item {\em It is a simple model:} With $d$ inputs, both the space and computational complexity are $O(d)$. It is a model that is easy to implement (e.g., on the ``edge'') and easy to parallelize if there is suitable hardware available.

\item {\em It does not overfit:} Being a simple model, it has little variance and can be trained on small data sets.

\item {\em It is easy to interpret:} Once a linear model has been trained, we can assess the influence of input features $x_j$ by looking at their associated weights $w_j$. Whether a weight is positive or negative tells us whether the corresponding feature is a supporting or inhibiting factor, and the magnitude of the weight tells us its strength. A feature whose associated weight is very close to zero can be removed, and a regularization term can be added to Equation~\ref{eq-linmod-err} to pull unnecessary weights towards zero, which implies that the multivariate linear model can have its built-in feature selection. 
\end{itemize}

Despite these various advantages, the model is limited to a linear approximation; that is, it can learn a linear regression, or can only be used when the classes are linearly separable. Thus, there may be bias if the regression/discriminant function that underlies the data is nonlinear.

The model, however, can be generalized by the use of {\em nonlinear basis functions}
\begin{equation}
y^t_i = f\left( \sum_{h=0}^H w_{ih} z^t_h \right) = f\left( \sum_{h=0}^H w_{ih} \phi_h(\bx^t) \right) , i=1,\ldots,K
\label{eq-genlinmod}
\end{equation}

\noindent where $z^t_h \equiv \phi_h(\bx^t)$ implement a mapping from the original $d$-dimensional $\bx$-space to the $H$-dimensional space of $z_h$. The linear model of Equation~\ref{eq-genlinmod} in this new $\bz$-space corresponds to a nonlinear model in the original $\bx$-space.

These basis functions can be fixed using prior information. One possibility is the higher-order terms of $\bx$. For example with $x\in\Re$, we can define $z_1=\phi_1(x)=x$ and $z_2=\phi_2(x)=x^2$, and $y=w_0 + w_1 z_1 + w_2 z_2$ is linear in $(z_1, z_2)$ but is quadratic in $x$: $y=w_0 + w_1 x + w_2 x^2$. One can use other basis functions as well, and in the past, people have proposed a variety of such basis functions, such as $\exp()$, $\sin()$, and so on, or transformations, for example, wavelets. Another possibility is to use an unsupervised method like principal components analysis to calculate the first-layer weights (which are orthogonal projection directions). Regardless of how, as long as the basis functions are fixed, Equation~\ref{eq-genlinmod}, though nonlinear in its input, is still a model that is linear in its parameters and can still be trained optimally.

From a neural network perspective, Equation~\ref{eq-genlinmod} is a multilayer perceptron (MLP) where $z_h$ are the hidden units and $w_{ih}$ are the second-layer weights. For this case, we have
\begin{equation}
z^t_h = \phi (\bx^t ; \br_h ) = g\left( \sum_{j=0}^d r_{hj} x^t_j \right) , h=1,\ldots,H
\label{eq-hiduni}
\end{equation}

\noindent where the advantage is that $\phi()$ are parameterized by $\br_h$ that can be trained from data. 

Each $z_h$ takes a weighted sum of the inputs and passes the sum through a nonlinear activation function $g()$, e.g., sigmoid, ReLU, and so on. As long as $g()$ is differentiable, one can use gradient descent on the error function of Equation~\ref{eq-linmod-err} to update both $w_{ih}$ {\em and\/} $r_{hj}$; this is called {\em back-propagation}. One can have multiple layers of such hidden units, each one taking a weighted sum of the units in the preceding layer and feeding its output to the units in the layer that follows; this is known as {\em deep learning}. 

The caveat is that with some of the parameters inside the nonlinear basis functions, the error function of Equation~\ref{eq-linmod-err} becomes non-convex and back-propagation converges to a local minimum. Another problem is that a network with multiple layers of modifiable weights has too many free parameters and may overfit the training data, failing to generalize. It is also known that gradient descent may be too slow on deep networks despite heuristics such as momentum factors and adaptive learning rates; when there are many layers, the gradients may vanish or explode. 

Another possibility is to keep Equation~\ref{eq-hiduni}, but set $r_{hj}$ to random values. This idea has particularly been investigated in the case of an MLP with a single hidden layer where the first-layer weights, $r_{hj}$, are randomly set and kept fixed, and only the second-layer weights, $w_{ih}$, are trained on data. 

What we propose is the {\em half-layer model} that is in between. We propose to define a half unit where $\br_h$ is decomposed into two stages of weights, where one stage has random weights and the other stage has trained weights. 

This paper is organized as follows: Our proposed model and how it differs from the fully trained and fully random models are discussed in Section \ref{sec:hhalf}.  We will discuss first the case with a single hidden layer but we will also consider the case where such randomized layers can be integrated into deep MLPs. The experimental results on MNIST and FashionMNIST data sets are given in Section \ref{sec:exp}. We discuss our findings and conclude in Section \ref{sec:conc}.

\section{Full vs. Half Layers}
\label{sec:hhalf}

\subsection{A Half Layer is Half-Fixed and Half-Trained}

In a vanilla MLP, both the first-layer weights $r_{hj}$ and the second-layer weights $w_{ih}$ are trained on data. The idea in some studies   \cite{rahimi07,huang12,cao17,scar17,gall20} is to fix $r_{hj}$ by assigning them random values, updating only $w_{ih}$ using data. The advantage of fixing the first layer is that the second layer, which contains the trained weights, is a linear model, resulting in a convex optimization problem.

In hidden units, we can interpret $\br_h$ as linear projection directions. The hidden units map $\bx$ to a new $H$-dimensional space, and if $\br_h$ are set at random, those projection directions are chosen at random. Using a large enough $H$, we may believe that just by luck we may hit on some useful directions; a suitable regularization in the second layer could help us get rid of the useless projections. 

But it seems that if we are to use a nonlinear $g()$ in Equation~\ref{eq-hiduni}, we have to be careful when we assign values to the weights $r_{hj}$ and the biases $r_{h0}$. For example, if $g()$ is the sigmoid, we have to make sure that we do not saturate it, or if it is ReLU, we make sure to stay on the positive side. The magnitudes of those first-layer weights should depend on the scale of $x^t_j$ as well as the input dimensionality $d$. So, there must be some ``under the hood'' preprocessing that should be done depending on the application, even if the weights are chosen at random.

\begin{figure}
  \centering
\includegraphics[width=0.8\textwidth]{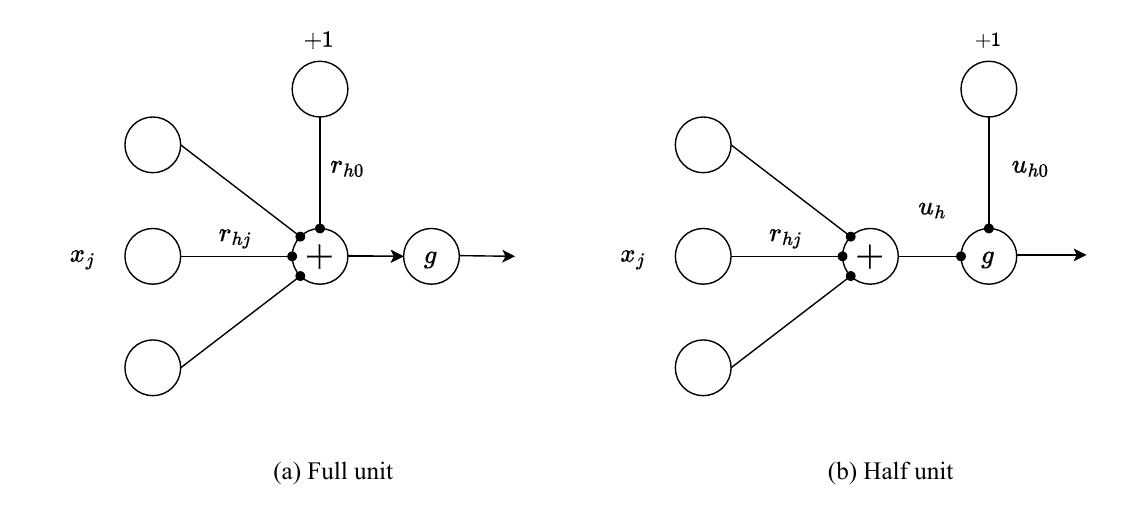}
\caption{(a) Full unit vs (b) half unit. In (a), all $r_{hj}$ are randomly set or they are all trained; in (b), $r_{hj}$ are randomly set whereas $u_h$ and $u_{h0}$ are trained.}
\label{fig-halfunit}
\end{figure}

Figure~\ref{fig-halfunit}(a) is a conventional hidden unit where the weights $r_{hj}$ are all trained or all randomly fixed. Our proposed half unit is shown in Figure~\ref{fig-halfunit}(b), where the processing in a hidden unit is decomposed into two stages:
\begin{eqnarray}
a^t_h 	&=& \sum_{j=1}^d r_{hj}x^t_j 
\label{eq-halfact}\\
z^t_h	&=& g( u_h a^t_h + u_{h0})	
\end{eqnarray}

\noindent where the first-stage weights $r_{hj}$, are fixed to random values and the second-stage weights, $u_h, u_{h0}$, are trained using back-propagation. 

The first stage calculates the total activation $a_h$ (by a dot product), and the second stage scales this by $u_h$ and translates by $u_{h0}$, before the result is passed through the nonlinear $g()$. Since only part of the parameters are updated, we call this a {\em half unit}, and a {\em half layer} is composed of such half units. 

$r_{hj}$ can be in any scale; $u_h$ and $u_{h0}$ learn to adapt the total activation to the scale of the particular data at hand. Note also that we have only two modifiable parameters per hidden unit, and so the number of modifiable parameters in such a layer is independent of the input dimensionality $d$. It is true that with modifiable $u_h, u_{h0}$ inside $g()$, the error function is non-convex, but we believe that with much fewer parameters, we have a simpler function that is easier to optimize. 

\subsection{Choosing the Weights of a Half Layer}

The question now is under which conditions adding a fixed random layer leads to an improvement? Or, in other words, how can we randomly choose the weights of the first layer so that $z^t_h$ becomes a better representation than the original $x^t_j$?

It seems at first too optimistic to hope to achieve a useful mapping by randomly chosen first-layer weights, but then we remember that even with back-propagation on a full MLP, the first-layer weights that we get are random approximations because of all the various sources of randomness in training: The training set is a random sample, the initial weights are random, there is stochasticity in sampling the instances/mini-batches, and so on. And we may knowingly add even further randomness, e.g., by dropout. 

\begin{figure}
\begin{center}
(a) Unit normal\\
\includegraphics[width=0.4\textwidth]{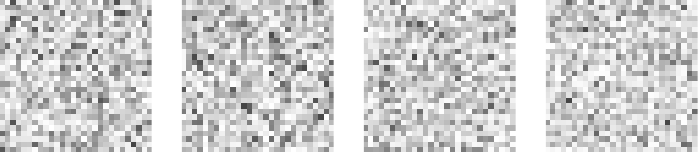}

(b) Binary \\
\includegraphics[width=0.4\textwidth]{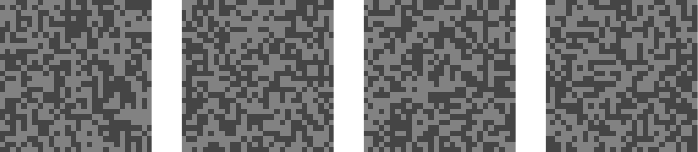}

(c) Random Mexican hats\\
\includegraphics[width=0.4\textwidth]{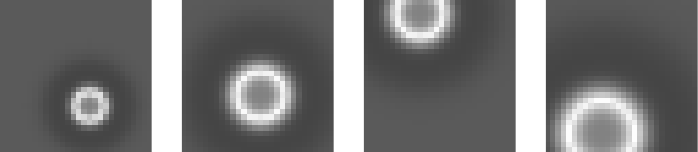}

(d) Random instances\\
\includegraphics[width=0.4\textwidth]{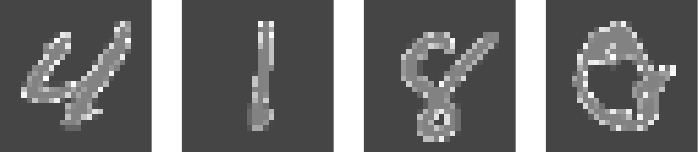}
\caption{Different ways of randomly setting the weights of a half layer.}
\label{fig:samples}
\end{center}
\end{figure}

We investigate several ways to set $r_{hj}$ randomly (see Figure \ref{fig:samples}):
\begin{enumerate}
\item {\em Random Unit-Normal Weights} (N): Each $r_{hj}, \forall h, j$ is drawn independently from $N(0,1)$. Because of the shape of the unit normal, most weights will be close to zero, implying a built-in regularization effect.

\item {\em Random Binary Weights} (B): Each $r_{hj}, \forall h, j$ is drawn independently from a Bernoulli distribution $\{-1,+1\}$ with $p=0.5$. This structure is interesting because (i) every weight is one bit (less memory), and (2) every multiplication of Equation~\ref{eq-halfact} is turned into a subtraction or addition (less calculation).

\item {\em Random Mexican Hats} (M): If the input is an image, each $\mathbf{r}_h$ can be set to be a Mexican-hat filter whose position and size are chosen randomly. Equation~\ref{eq-halfact} then is a convolution and acts like a blob detector. A layer of such units can be interpreted as a randomly constructed convolutional layer. One can envisage other similar, randomly parameterized filters, e.g., Gabor filters, or one-dimensional variants. 

\item {\em Random Instances} (T): Each $\mathbf{r}_h$ is set to a randomly drawn training instance, and so each $z_h$ works as a template-matcher. This is like the radial-basis function network or the Gaussian kernel in support vector machines, except that we use the dot product instead of the Euclidean distance.
\end{enumerate}

\section{Experimental Results}
\label{sec:exp}

We do experiments on two data sets. Both MNIST and FashionMNIST contain $28\times 28$ gray-scale images from ten classes with 60,000 training and 10,000 test examples. In MNIST, the ten classes are digits, and in FashionMNIST, they are fashion articles. Given that the two data sets have the same characteristics, we use the same network architectures on both.

All networks are trained five times using the same training and test data but with different random seeds; the reported values are the average and standard deviations of those five runs.

\subsection{Networks with One Hidden Layer}

As baselines, we have the following models: {\tt lp} is a one-layer network that uses the original 784 inputs; it has $(784+1)\times 10$ weights. {\tt mlp-H} is a multi-layer perceptron with one hidden layer of $H$ hidden units; it has $(784+1)\times H + (H+1)\times 10$ weights. Both are full-layered networks, i.e., all their weights are trained using back-propagation as usual. 

We denote our network with one hidden layer of $H$ half units as {\tt rnd-H-C} where $C$ is one of \{N, B, M, T\} as explained above. Such a network has $2\times H + (H+1)\times 10$ weights, all trained using back-propagation. Note again that the number of modifiable weights of such a network does not depend on the input dimensionality $d$, here 784; but of course, the calculation does depend on $d$; the first stage of a half unit (Equation~\ref{eq-halfact}) does not involve any modifiable parameter, but it still needs to be evaluated for all hidden units. 

\begin{figure}
\begin{center}
\includegraphics[width=0.6\textwidth]{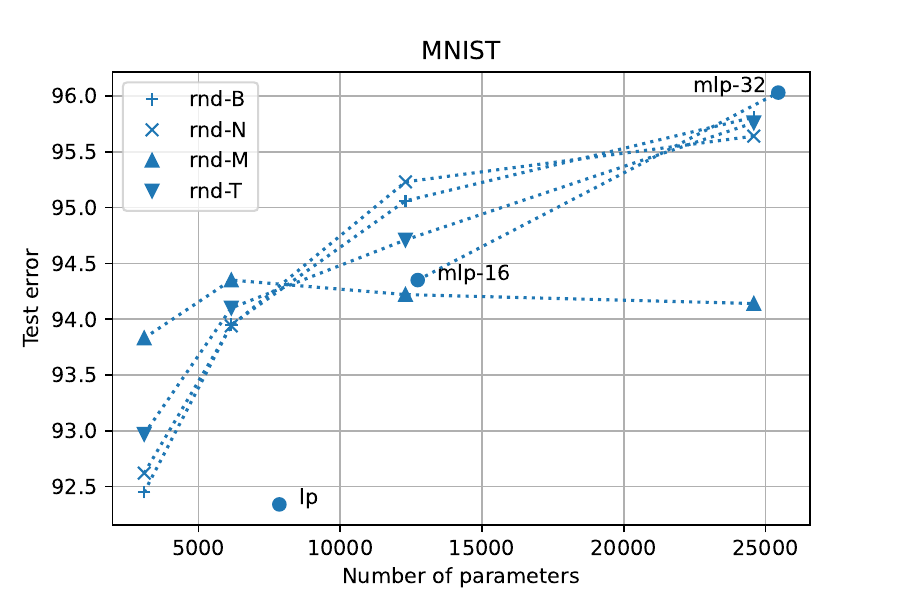} 
\caption{Results on MNIST with networks having one full or half hidden layer.}
\label{fig-mnist1}
\end{center}
\end{figure}

The results with all the networks on MNIST are given in Table~\ref{tab-mnist1} and plotted in Figure~\ref{fig-mnist1} in the two dimensions of the number of modifiable parameters and the average test accuracy. We see that the accuracy increases as we go from {\tt lp} to {\tt mlp} (having a hidden layer helps) and increases further with more hidden units.

We see that our proposed {\tt rnd} with 256 hidden units achieves higher test accuracy than {\tt lp}, despite having fewer modifiable parameters. This is the advantage of having a hidden layer, even though most of its weights are random. {\tt rnd} variants with 1024 random units have as many modifiable parameters as vanilla {\tt mlp} with 16 hidden units and are as accurate or better. Similarly, {\tt rnd} variants with 2048 units are as complex and as accurate as {\tt mlp-32}. 

\begin{figure}
\begin{center}
\includegraphics[width=0.6\textwidth]{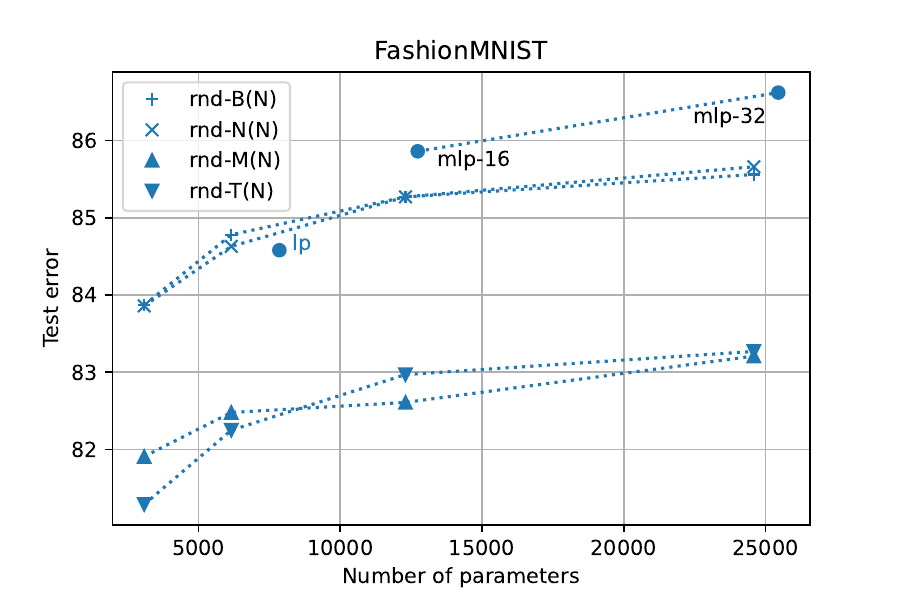}
\caption{Results on FashionMNIST with networks having one full or half hidden layer.}
\label{fig:fashionmnist1}
\end{center}
\end{figure}

The results on FashionMNIST, as given in Table~\ref{tab-fashionmnist1} and plotted in Figure~\ref{fig:fashionmnist1}, are similar. Again, we see that {\tt rnd} variants are more accurate than {\tt lp}, but slightly less accurate than {\tt mlp-16} or {\tt mlp-32} given enough random units. Just as we see on MNIST, the gain through random hidden units is not as significant when too many of them are used. This indicates that randomness only helps up to a certain level.

Among the {\tt rnd} variants, the M and T variants do not seem to work well on FashionMNIST. B and N variants seem to perform the best overall; this is especially promising given that B uses one-bit weights.

\begin{table}
\caption{Results of networks with one full or half hidden layer on MNIST.}\label{tab-mnist1}
\begin{center}
\begin{tabular}{lrrr}
\toprule
Model       & \# parameters 	& Training accuracy & Test accuracy\\
\midrule
lp      	& 7,850     	& 92.95, 0.01   & 92.34, 0.03 \\
\hline
mlp-16  	& 12,730    	& 96.24, 0.18   & 94.35, 0.08 \\
mlp-32  	& 25,450    	& 98.58, 0.16   & 96.03, 0.17 \\
\hline
rnd-256-N   & \multirow{4}{*}{3,082} & 93.01, 0.33   & 92.62, 0.34 \\
rnd-256-B   & 			& 92.83, 0.22   & 92.45, 0.39 \\
rnd-256-M   & 			& 94.38, 0.32   & 93.83, 0.27 \\
rnd-256-T   & 			& 93.79, 0.38   & 92.97, 0.30 \\
\hline
rnd-512-N   & \multirow{4}{*}{6,154} & 94.87, 0.22   & 93.94, 0.12 \\
rnd-512-B   & 			& 94.96, 0.30   & 93.95, 0.40 \\
rnd-512-M   & 			& 95.54, 0.37   & 94.35, 0.29 \\
rnd-512-T   & 			& 96.10, 0.32   & 94.10, 0.23 \\
\hline
rnd-1024-N   & \multirow{4}{*}{12,298} & 96.67, 0.15   & 95.23, 0.16 \\
rnd-1024-B   & 			& 96.30, 0.38   & 95.06, 0.37 \\
rnd-1024-M   &			& 95.96, 0.31   & 94.22, 0.30 \\
rnd-1024-T   & 			& 98.10, 0.19   & 94.71, 0.22 \\
\hline
rnd-2048-N   & \multirow{4}{*}{24,586} & 97.33, 0.21   & 95.64, 0.11 \\
rnd-2048-B   & 			& 97.54, 0.32   & 95.81, 0.35 \\
rnd-2048-M   & 			& 96.26, 0.51   & 94.14, 0.50 \\
rnd-2048-T   & 			& 99.27, 0.19   & 95.76, 0.08 \\
\bottomrule
\end{tabular}
\end{center}
\end{table}

\begin{table}
\caption{Results of networks with one full or half hidden layer on FashionMNIST}\label{tab-fashionmnist1}
\begin{center}
\begin{tabular}{lrrr}
\toprule
Model       & \# parameters 	& Training accuracy & Test accuracy\\
\midrule
lp      	& 7,850     & 86.78, 0.01   & 84.58, 0.02 \\
\hline
mlp-32  	& 12,730    & 89.06, 0.14   & 85.86, 0.24 \\
mlp-32  	& 25,450    & 90.43, 0.16   & 86.62, 0.29 \\
\hline
rnd-256-N   & \multirow{4}{*}{3,082} & 85.61, 0.16   & 83.86, 0.23 \\
rnd-256-B   & 			& 85.65, 0.17   & 83.87, 0.19 \\
rnd-256-M   & 			& 83.62, 0.46   & 81.91, 0.24 \\
rnd-256-T   & 			& 83.10, 0.56   & 81.29, 0.40 \\
\hline
rnd-512-N   & \multirow{4}{*}{6,154}	 & 86.87, 0.23   & 84.63, 0.24 \\
rnd-512-B   & 			& 87.08, 0.24   & 84.78, 0.30 \\
rnd-512-M   & 			& 84.55, 0.21   & 82.48, 0.30 \\
rnd-512-T   & 			& 84.64, 0.59   & 82.25, 0.62 \\
\hline
rnd-1024-N   & \multirow{4}{*}{12,298} 	& 87.91, 0.15   & 85.27, 0.20 \\
rnd-1024-B   & 			& 87.92, 0.19   & 85.27, 0.31 \\
rnd-1024-M   & 			& 84.89, 0.27   & 82.61, 0.31 \\
rnd-1024-B   & 			& 86.02, 0.35   & 82.97, 0.20 \\
\hline
rnd-2048-N   & \multirow{4}{*}{24,586} & 88.74, 0.26   & 85.66, 0.23 \\
rnd-2048-B   & 			& 88.67, 0.24   & 85.56, 0.35 \\
rnd-2048-M   & 			& 85.38, 0.23   & 83.21, 0.26 \\
rnd-2048-T   & 			& 86.52, 0.63   & 83.27, 0.57 \\
\bottomrule
\end{tabular}
\end{center}
\end{table}

\subsection{Convolutional Networks}

It is also possible to use such a half layer after convolutional layers. The error at the output can be back-propagated through half units to update the parameters of earlier convolutional layers.

We use a convolutional structure with two convolutional hidden layers; the first layer takes the $28\times 28$ image as input and has 16 ReLU $5\times 5$ filters followed by a $2\times 2$ pooling; the second layer has 32 ReLU $3\times 3$ filters followed by $2\times 2$ pooling. These convolutional layers use a total of $(16+1)\times 5\times 5 + (16\times 3\times 3 + 1)\times 32 = 5,856$ weights. The resulting output is flattened into a 1,152-dimensional vector.

In {\tt conv}, this output is fully connected to the ten output units with $(1152+1)\times 10=11,530$ weights. In {\tt conv2}, there is a fully connected hidden layer of 40 units between the two layers of 1,152 and 10 units. This has $(1152+1)\times 40 + (40+1)\times 10 = 46,530$ weights. These are full-layered networks, and all their weights are trained using back-propagation as usual. 

The convolutional version of our method, which we name {\tt crnd-H}, keeps the same convolutional structure and adds a half layer of $H$ units between the two layers of 1,152 and 10 units; this adds $2\times H + (H+1)\times 10$ weights. The parameters of the convolutional layers are trained as usual, except that the error is back-propagated through the half layer. Since most of the weights of a deep neural network lie not in the convolutional layers but the fully-connected ones, such layers seem particularly suitable candidates to be replaced by half layers.

\begin{figure}
\begin{center}
\includegraphics[width=0.6\textwidth]{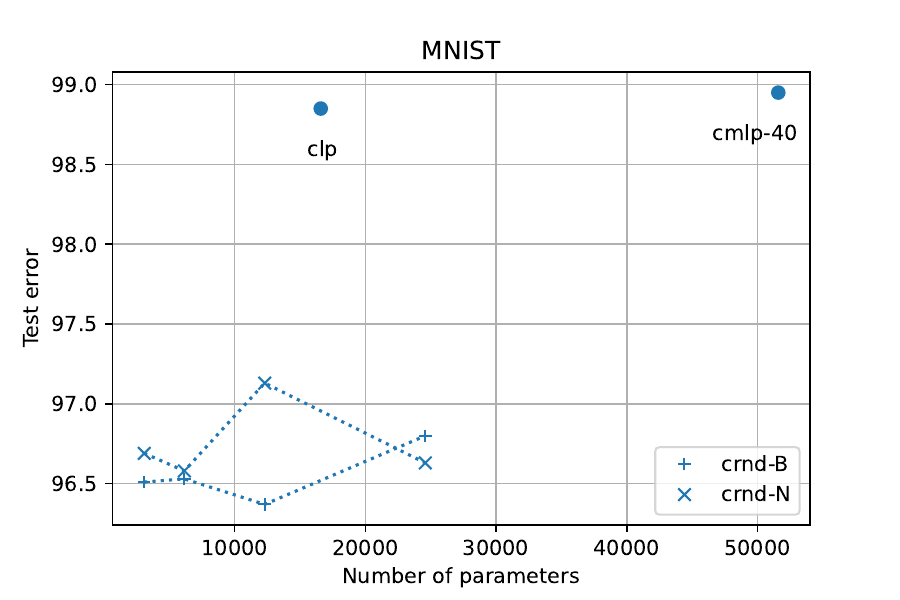}  
\caption{Convolutional network results on MNIST.}
\label{fig:mnist2}
\end{center}
\end{figure}

In Table~\ref{tab-mnist-conv} and Figure~\ref{fig:mnist2}, we see the results on MNIST. With the convolutional structure, the accuracies of {\tt rnd} variants increase, but not as much as with the fully-trained {\tt mlp} variants. On FashionMNIST (see Table~\ref{tab-mnist-conv} and Figure~\ref{fig:mnist2}), we see that adding convolutional layers helps with vanilla {\tt mlp}, though not with {\tt rnd}. There does not seem to be a significant difference in accuracy between the B and N variants (M and T variants are not applicable here).

\begin{table}
\caption{Results of convolutional network variants on MNIST}\label{tab-mnist-conv}
\begin{center}
\begin{tabular}{lrrr}
\toprule
Model       & \# parameters 	& Training accuracy & Test accuracy\\
\midrule
clp     & 16,586    & 99.45, 0.05   & 98.85, 0.03 \\
cmlp-40 & 51,586    & 99.65, 0.05   & 98.95, 0.09 \\  
\hline
crnd-256-N   & \multirow{2}{*}{8,138} & 96.61, 0.31   & 96.69, 0.45 \\
crnd-256-B   &			& 96.52, 0.59   & 96.51, 0.59 \\
\hline
crnd-512-N   & \multirow{2}{*}{11,210} & 96.66, 1.07   & 96.58, 1.01 \\
crnd-512-B   & 			& 96.48, 0.47   & 96.53, 0.47 \\
\hline
crnd-1024-N   & \multirow{2}{*}{17,354} & 97.18, 0.45   & 97.13, 0.43 \\
crnd-1024-B   & 		& 96.43, 0.89   & 96.37, 0.91 \\
\hline
crnd-2048-N   & \multirow{2}{*}{29,642} & 96.75, 0.57   & 96.63, 0.65 \\
crnd-2048-B   & 		& 96.86, 0.67   & 96.80, 0.68 \\
\hline
\end{tabular}
\end{center}
\end{table}

\begin{table}
\caption{Results of convolutional networks on FashionMNIST}\label{tab-fashionmnist-conv}
\begin{center}
\begin{tabular}{lrrr}
\toprule
Model       & \# parameters 	& Training accuracy & Test accuracy\\
\midrule
clp      & 16,586    & 92.76, 0.17   & 90.32, 0.17 \\
cmlp-40  & 51,586    & 92.77, 0.38   & 89.37, 0.30 \\  
\hline
crnd-256-N   & \multirow{2}{*}{8,138} & 86.27, 0.35  & 85.12, 0.41 \\
crnd-256-B   & 			& 85.95, 0.25  & 84.89, 0.25\\
\hline
crnd-512-N   & \multirow{2}{*}{11,210} & 86.53, 0.37   & 85.37, 0.54 \\
crnd-512-B   & 			& 86.61, 0.67   & 85.64, 0.70 \\
\hline
crnd-1024-N   & \multirow{2}{*}{17,354} & 86.95, 0.57   & 85.77, 0.66 \\
crnd-1024-B   & 		& 87.22, 0.56   & 86.16, 0.60 \\
\hline
crnd-2048-N   & \multirow{2}{*}{29,642} & 89.90, 1.80   & 84.08, 1.08 \\
crnd-2048-B   & 		& 85.53, 0.45   & 84.45, 0.42 \\
\hline
\end{tabular}
\end{center}
\end{table}

\begin{figure}
\begin{center}
\includegraphics[width=0.6\textwidth]{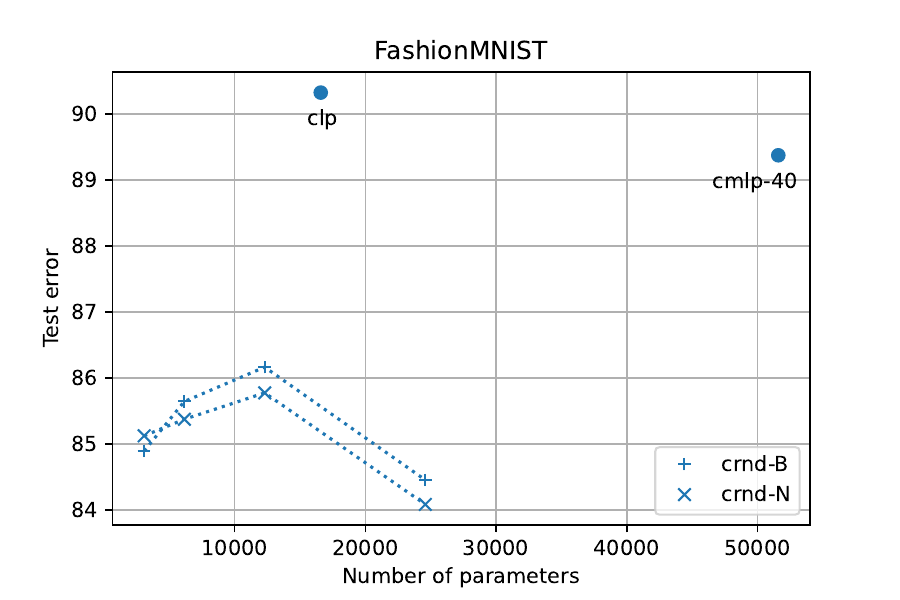} 
\caption{Convolutional network results on FashionMNIST.}
\label{fig:fashionmnist2}
\end{center}
\end{figure}

\begin{figure}
\begin{center}
\begin{tabular}{cc}
\includegraphics[width=0.45\textwidth]{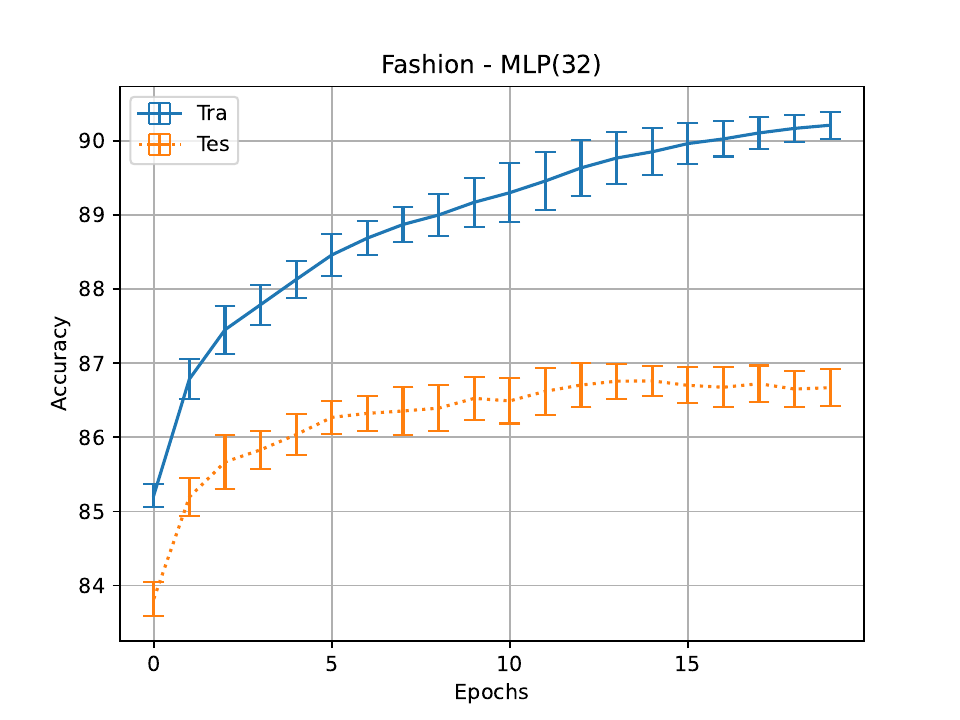} &
\includegraphics[width=0.5\textwidth]{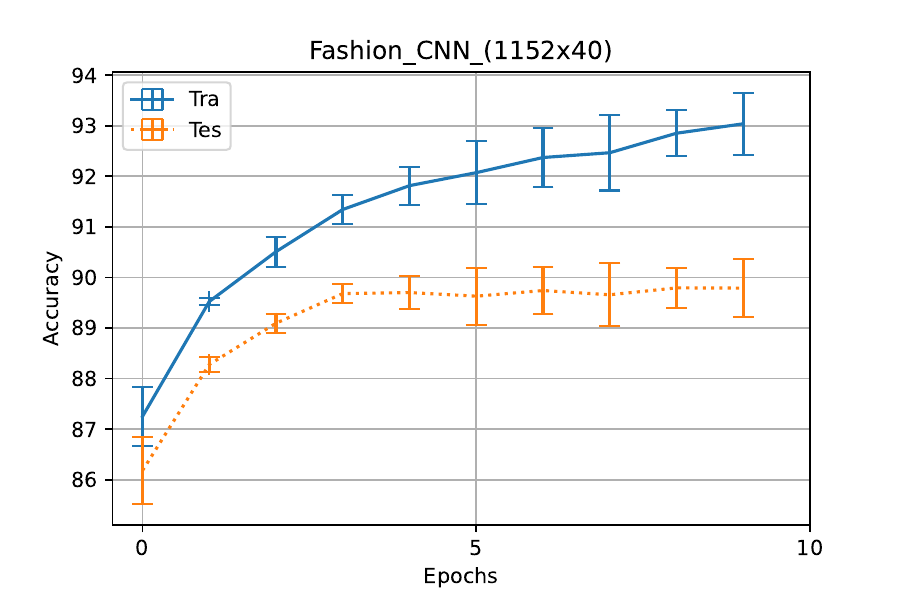}
\end{tabular}
\caption{Learning curves of (a) {\tt mlp32} and (b) {\tt conv2} on FashionMNIST.}
\label{fig:fashionmnistcomp1}
\end{center}
\end{figure}

To help us understand this difference in behavior, we compare the learning curves of some example {\tt mlp} and {\tt rnd} variants on FashionMNIST. In Figure~\ref{fig:fashionmnistcomp1}, we look at those of {\tt mlp32} and {\tt conv2}; the first has one hidden layer, and the second is convolutional. Both are full-layered networks, and on both, we see that there is a clear distinction between the training and test accuracies during training. Beyond a certain number of epochs, training accuracy continues to increase, but test accuracy does not, which is a sign of overfitting.

\begin{figure}
\begin{center}
\begin{tabular}{cc}
\includegraphics[width=0.5\textwidth]{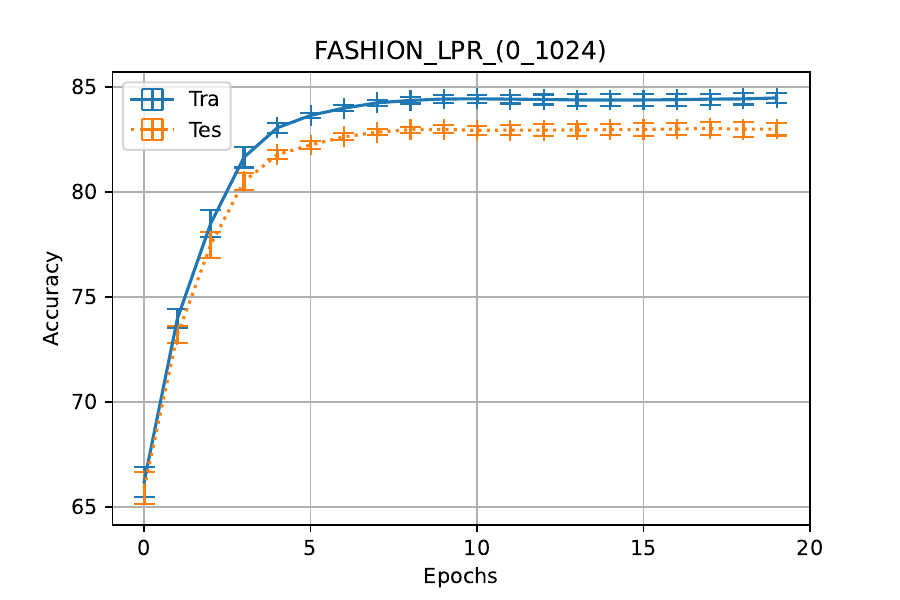} &
\includegraphics[width=0.5\textwidth]{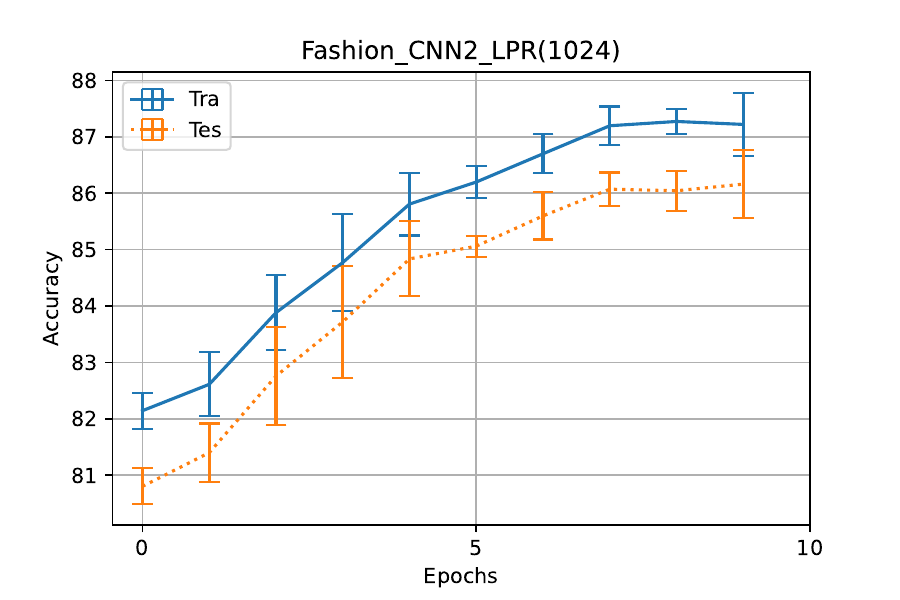}
\end{tabular}
\caption{Learning curves of (a) {\tt rnd-B-1024} and (b) {\tt convrnd-B-1024} on FashionMNIST.}
\label{fig:fashionmnistcomp2}
\end{center}
\end{figure}

In contrast, in Figure~\ref{fig:fashionmnistcomp2}, we see the learning curves of two {\tt rnd} variants: The first one has one hidden layer of 1024 half units fed directly with the input, and the second one has convolutional layers before 1024 half units. On both, we see that the training and test accuracies move together during training. We believe this to be due to the smoothing (regularizing) effect of the random units. Many randomized units that act together avoid overfitting, but the disadvantage is that in certain cases, as we have here on FashionMNIST, they smooth too much, introducing bias. We believe this to be the reason why the accuracy is not always as high, and as we see in Figure~\ref{fig:mnist2} when going from 1024 to 2048, accuracy may even get worse with too many random units. 

\section{Conclusions}
\label{sec:conc}

We investigate an MLP structure where a hidden unit has some of its parameters randomly set and fixed, and some of them are trained on data. We show that a layer of such half units may be used in the first or in a later layer of a deep network. Because the number of trained parameters does not depend on the fan-in, we believe that such structures are especially interesting as fully-connected layers, which, in general, make the greatest contribution to the total number of free parameters in a deep neural network.

Our experimental results indicate that such half layers may achieve accuracies as high as networks with full layers, but not always. We can view each half unit as a noisy base-learner, and a layer of them working together may be seen as an ensemble model. The averaging/voting effect helps with decreasing variance, but may also introduce bias if the smoothing effect of the randomness is more than the bias-reducing effect of the trained components.

These findings are the result of experiments on relatively simple networks on relatively simple datasets, and further experimentation is needed on larger datasets with deeper networks for more general conclusions.

\bibliographystyle{ACM-Reference-Format}
\bibliography{halfNN}

\end{document}